\documentclass[letterpaper, 10 pt, conference]{ieeeconf}  %
\IEEEoverridecommandlockouts                              
\usepackage{graphicx} 
\usepackage{amsmath} 
\usepackage{amssymb}  
\usepackage{booktabs}
\usepackage{cite}
\usepackage{lipsum}
\usepackage{comment}
\usepackage{flexisym}
\usepackage{amsmath}
\usepackage{subcaption}
\usepackage{flushend}
\usepackage{algpseudocode}
\usepackage{algorithm}
\usepackage{lipsum}
\usepackage{bm}

\usepackage{svg}

\algnewcommand\algorithmicforeach{\textbf{for each}}
\algdef{S}[FOR]{ForEach}[1]{\algorithmicforeach\ #1\ \algorithmicdo}

\usepackage{subcaption}
\usepackage{cite}
\usepackage[normalem]{ulem}

\title{Online Monitoring of Object Detection Performance During Deployment} 
\author{Quazi Marufur Rahman, Niko S{\"u}nderhauf and Feras Dayoub %
\thanks{The authors are with the Australian Centre for Robotic Vision at Queensland University of Technology (QUT),
Brisbane, QLD 4001, Australia. This research has been conducted by the  Australian Research Council (ARC) Centre of Excellence for Robotic Vision (Grant CE140100016). The authors acknowledge continued support from the QUT Centre for Robotics. Contact: {\tt\small quazi.rahman@qut.edu.au}}%
}

\begin{document}

\maketitle

\begin{abstract}

During deployment, an object detector is expected to operate at a similar performance level reported on its testing dataset.  However, when deployed onboard mobile robots that operate under varying and complex environmental conditions, the detector's performance can fluctuate and occasionally degrade severely without warning. Undetected, this can lead the robot to take unsafe and risky actions based on low-quality and unreliable object detections. We address this problem and introduce a cascaded neural network that monitors the performance of the object detector by predicting the quality of its mean average precision (mAP) on a sliding window of the input frames. The proposed cascaded network exploits the internal features from the deep neural network of the object detector. We evaluate our proposed approach using different combinations of autonomous driving datasets and object detectors.

\end{abstract}

\section{Introduction}

\begin{figure}[pt]
 \centering
 \includegraphics[width=0.99\columnwidth]{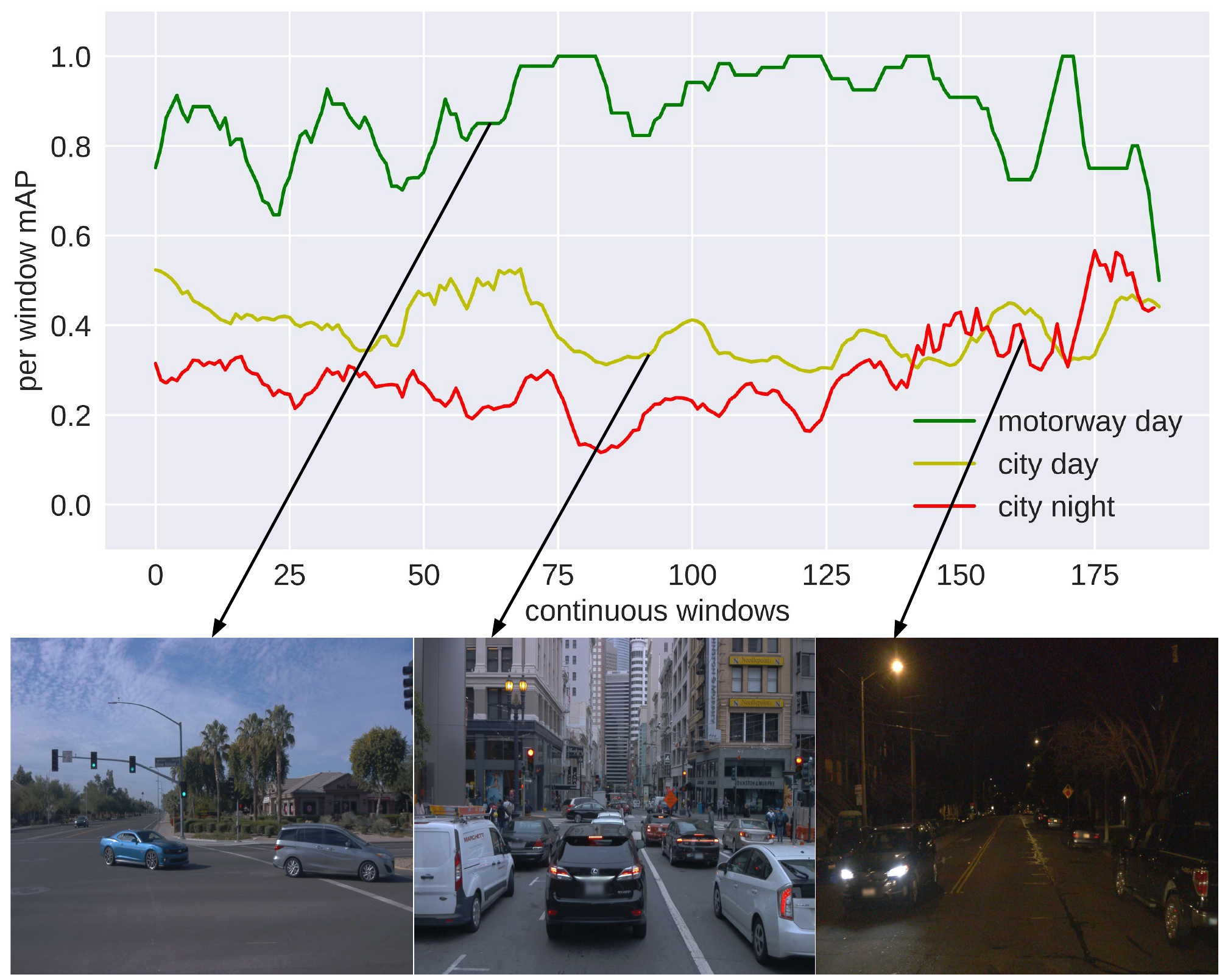}
 \caption{Object detector performance deployed on a self-driving car depends heavily on multiple factors like traffic density, road type, and time of the day. The fluctuation in the mAP (calculated over a sliding window of 10 frames) shows the performance instability during the deployment phase in multiple places and conditions. We show that a specialized performance monitoring network can predict the mAP of the object detector to inform downstream tasks of its expected reliability. In this figure, the $1^{st}$ row shows the continuous per-window mAP of three road scenes (motorway day, city day and night). The $2^{nd}$ row shows one sample image from each road scene. }
\label{fig:fig1}
\end{figure}

Object detection plays a vital role in many robotics and autonomous system applications. For instance, a driverless car is expected to detect important objects such as vehicles, people and traffic signs accurately all the time.  Failure to do so can cause severe consequences for the car and the people involved. Hence, there is ongoing research \cite{tian2019fcos, ren2015faster, liu2016ssd, Duan2019CenterNetKT, Bochkovskiy2020YOLOv4OS, He2019RethinkingIP, Cai2018CascadeRD, Li2018DetNetAB, Lin2017FocalLF, Carion2020EndtoEndOD} to improve the robustness and accuracy of object detection systems. In general, an object detection system is trained and evaluated using non-overlapping training, validation, and test splits of a dataset before deployment. The underlying assumption is that the images encountered during deployment follow a similar distribution to the images presented before deployment. However, in the case of autonomous systems, the deployment environment can exhibit many conditions that are not well represented in the training and evaluation datasets. This leads to the fact that during the deployment phase, performance can fluctuate and may diverge from the expected training and evaluation phase performance without any prior warning. Such silent change in performance during the deployment phase is a serious concern for any vision-based robotic system, see Fig.~\ref{fig:fig1}. 
 
The ultimate solution to meet this challenge is to develop a remarkably persistent object detection system by collecting training data from all imaginable conditions that can be encountered during the deployment phase. As such a solution is not practical, one remedy to this situation is to deploy a performance monitoring system for the object detector that can raise warnings when the performance drops below a critical threshold. 

A performance monitoring system is expected to provide the capability of self-assessment to the object detector. This self-assessment can improve safety and robustness during the deployment by monitoring the performance continuously and allowing to take preventive measures when the performance drops below the expected level. 

To this end, the contribution of this paper is a novel cascaded neural network that exploits the internal feature maps from the deep neural network of the object detector for the task of online performance monitoring. Our proposed cascaded network operates on a sliding window of frames and continuously predicts the performance of the object detector in terms of mean average precision (mAP). We evaluate our proposed approach against multiple baselines using different combinations of datasets and object detection networks.

The rest of the paper is organized as follows: In Section~\ref{sec:literature review}, we review the related works on performance monitoring. In Section~\ref{sec:approach overview}, we introduce our method for online performance monitoring of object detection during the deployment phase. Section~\ref{sec:experimental setup} outlines our experimental setup. Section~\ref{sec: evaluation and results} presents the results and finally in Section~\ref{sec:conclusion} we draw conclusion for this work.

\section{Related Works}
\label{sec:literature review}

Self-assessment and performance monitoring in robotics applications are important due to the high requirements of safety and robustness. In~\cite{morris2007robotic}, a framework called robotic introspection is developed to provide a self-assessment mechanism for field robots during exploration and mapping of subterranean environments. Later \cite{grimmett2016introspective} and \cite{Triebel2013DrivenLF} extended this work for obstacle avoidance and semantic mapping assessment. These works examine the output of the underlying models to predict their expected performance.  

Another approach to address the performance monitoring problem is to evaluate model input before inference. \cite{zhang2014predicting} proposed a framework following this paradigm. They train an alert module to find cases where the target model will fail. Later, a similar approach was used for failure prediction for MAV \cite{daftry2016introspective}, hardness predictor \cite{wang2018towards} for image classifiers, and probabilistic performance monitoring for robot perception system \cite{Gurau2018LearnFE} for the task of pedestrian detection based on past experience from repeated visits to the same location.  

Exploiting model confidence and uncertainty is another line of research to monitor the performance of a target model. Trust score \cite{Jiang2018ToTO}, maximum class probability \cite{Hendrycks2017ABF} and true class probability \cite{corbiere2019addressing} are some recent works based on model confidence to identify the failure of an underlying image classifier. In the context of uncertainty estimation,  \cite{Gal2016DropoutAA} proposed to use dropout as a Bayesian approximation technique to represent model uncertainty. Later, \cite{Devries2018LeveragingUE, Huang2018EfficientUE} applied this idea to identify the quality of image and video segmentation network. Out-of-distribution (OOD) detection \cite{hendrycks2016baseline, liang2017enhancing, hendrycks2018deep, lee2017training, lee2018simple} is another relevant field of research that can be useful for performance monitoring. While OOD detection focuses on identifying previously unknown input to a model, performance monitoring emphasizes identifying model accuracy for each-and-every input.

In the object detection context, there are few works that address the performance monitoring to some extent. \cite{Miller2018DropoutSF} and \cite{Cheng2018DecoupledCR} use dropout sampling and hard false positive mining, respectively, to identify object detection failures. \cite{Rahman2019DidYM} and \cite{Ramanagopal2018FailingTL} use internal and handcrafted features of an object detector respectively to identify false negative instances during deployment. These works focus on a per-object and per-frame basis and do not provide an overall assessment of the object detector performance considering the combined aspects of false positives, false negatives and object localization accuracy. These aspects are captured by a summary metric such as mAP. Our proposed approach can monitor object detection performance online by predicting the quality of its mAP for a sequence of images during the deployment phase without using any ground-truth data.

\section{Approach Overview}
\label{sec:approach overview}

In this section, we present our approach to online monitoring of object detection performance during the deployment phase. We start by formalizing the problem. Then we describe our proposed cascaded neural network architecture that operates on the feature stream generated by the underlying object detection network to monitor its performance in real-time. 

Let us denote an object detection network as  $od_{net}$, which is deployed on a driverless car to detect objects of interest like vehicle and pedestrian from the road. It takes a continuous stream of images $I = \{I_{1}, I_{2}, \ldots , I_{N}\}$ and detects all possible objects from each $I_{i}$. Our goal is to monitor the performance of $od_{net}$ by predicting its mAP continuously over a sliding window of images. 

As described by \cite{azulay2018deep}, modern deep CNN's become unstable when the input image is translated, re-scaled or slightly transformed by any other means. This observation holds for object detectors deployed on a driverless car too, where the mAP between two consecutive frames might vary significantly because of irrelevant and negligible changes in the viewpoint. As a result, per-frame performance monitoring can raise unnecessary false alarms. To mitigate this issue, we are adopting per sliding window performance monitoring, where the mAP between two consecutive windows does not change drastically. Hence, the performance monitoring network is expected to produce a consistent prediction by examining a sequence of images. To achieve this, we will deploy a second convolutional neural network that will access the internal features of $od_{net}$ to predict the quality of the mAP for each sliding window of images. This second network will be referred to as the performance monitoring network, $pm_{net}$.

Instead of processing each input image $I_{i}$ like $od_{net}$ does at a time, $pm_{net}$ operates on $\bm{\omega}$ sequential images and predict the overall mAP of $od_{net}$ on these $\bm{\omega}$ images. We will use $\bm{\omega}$ to refer to the window size used by $pm_{net}$. Here, $pm_{net}$ takes a stream of windows $W = \{W_{1}, W_{2}, \ldots , W_{M}\}$ and monitor the performance for each $W_{j}$, where $W_{j} = \{I_{i}, I_{i+1}, \ldots , I_{i+w}\}$. 

We formulate the task of performance monitoring as a multi-class classification problem consisting of $C$ classes. To do so, the per-window mAP range is split into $C$ equal and consecutive parts and labelled from $0$ to $C-1$.  We will denote these per-window mAP labels using $mAP_{w}$. The lowest and the highest label, $0$ and $C-1$, refer to the worst and the best possible classes, respectively.  As there is an ordinal relation among these labels, we will consider this multi-class classification problem as ordinal classification \cite{cardoso2007learning}.

Our proposed $pm_{net}$ exploits the features generated by $od_{net}$ during per frame inference. $od_{net}$ uses a backbone architecture $B$ to extract features for the inference task where $B$ is a collection of interconnected convolutional layers. During the inference, for each image $I_{i}$, $B$ generates a set of $p$ feature maps, $L_{i} = \{L_{i1}, L_{i2}, \ldots , L_{ip}\}$. Here, shape of $L_{ij}$ is $c_{ij} \times h_{ij} \times w_{ij}$. $c_{ij}$, $w_{ij}$ and $h_{ij}$ are the channel, width and height of the $j^{th}$ convolutional layer of the $i^{th}$ image.

After each inference $pm_{net}$ extracts $L_{i}$ from $B$ for input $I_{i}$ and applies channel-wise average pooling to convert each $3D$ features into $2D$. Now the converted set of feature is $\bar L_{i} = \{\bar L_{i1}, \bar L_{i2}, \ldots , \bar L_{ip}\}$ and shape of $\bar L_{ij}$ is $1 \times h_{ij} \times w_{ij}$. These operations are performed for all $I_{i}$ and the newly formed corresponding $2D$ features are stacked together in channel-wise direction. That means the $\bar L_{(i+1)j}$ is stacked with $\bar L{ij}$. After processing $\bm{\omega}$ images we get the feature $F_{W_{i}}$ for $W_{i}$. Here $F_{W_{i}} = \{F_{1}, F{2}, \ldots, F_{p} \}$ and $F_{i}$ has the size of $\bm{\omega}  \times h_{ij} \times w_{ij}$. The task of $pm_{net}$ is to predict $mAP_{w}$ from $F_{W}$.

We design the $pm_{net}$ as a cascaded convolutional neural network to train it to predict $mAP_{w}$ from $F_W$. Here, each layer of $pm_{net}$ is implicitly connected with all the previous layers through their individual convolutional filter. Using this network, we exploit the rich multi-level semantic features generated by the $od_{net}$ instead of only using the last convolutional layer features. $pm_{net}$ uses a set of convolutional filters $f = \{f_{1}, f_{2}, \ldots, f_{p-1}\}$ to propagate the features of $F_{W}$ from one layer to the next. Each filter $f_{i}$ operates on $F_{i}$ to generate a new feature $\bar F_{i}$ which has the same shape of $F_{i+1}$. Now a concatenation is performed to join $\bar F_{i}$ and $F_{i+1}$ in channel-wise direction. This set of operations can be formulated using Equation~\ref{eqn:conv_layers}.
\begin{equation}
\label{eqn:conv_layers}
    \mathcal{F} = f_{i}(F_{i}) \oplus F_{i+1}; \quad i=1,2,\ldots,p-1
\end{equation}

\begin{figure}[t]
\centering
\centering
    \includegraphics[width=0.99\columnwidth]{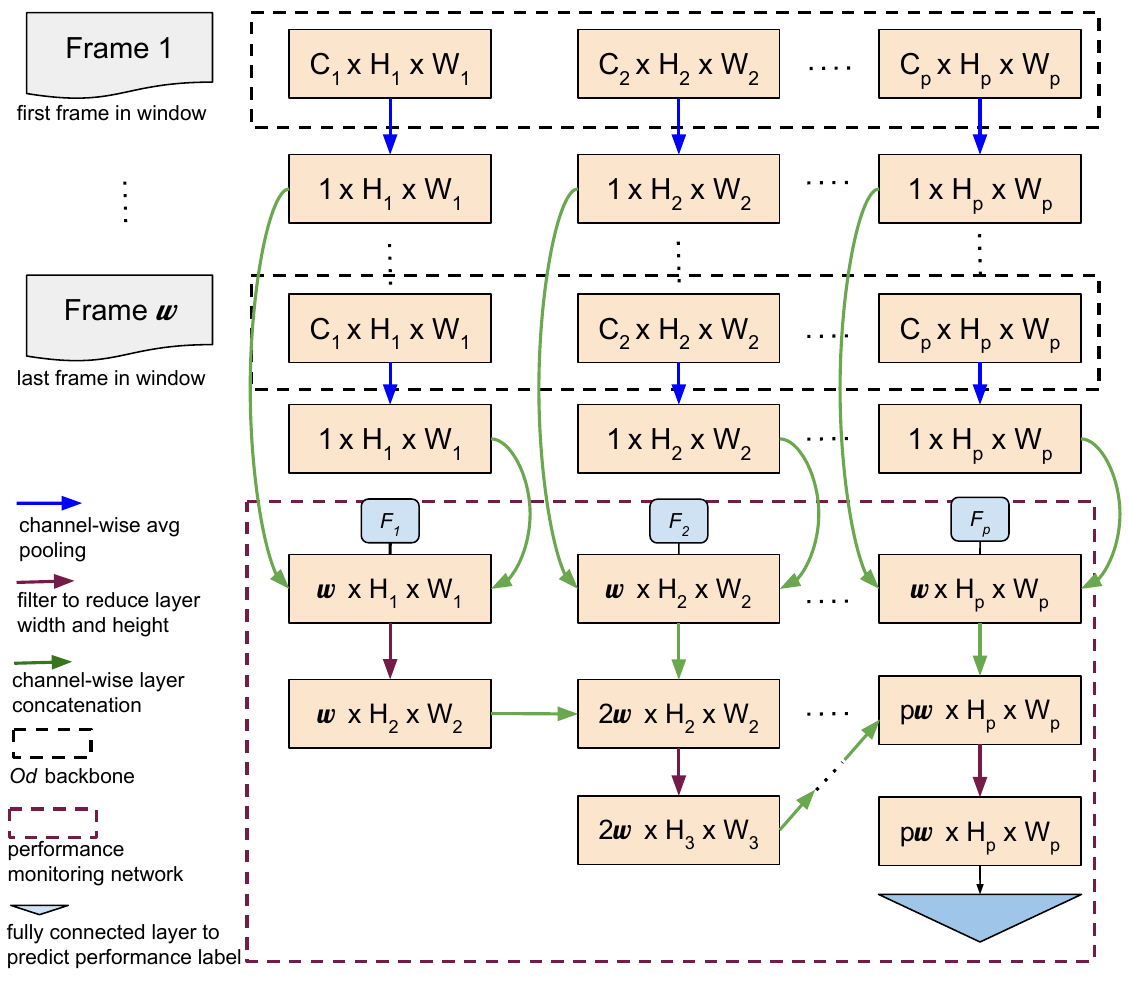}
\caption{Cascaded architecture of the proposed performance monitoring network. The first four rows show the procedure to convert the $3D$ feature from frame 1 to $\bm{\omega}$ into $2D$. These features are collected from the object detection backbone. $5^{th}$ row demonstrates how the $2D$ features from corresponding layers are stacked together to generate a $\bm{\omega}$ channel $3D$ feature. $6^{th}$ and $7^{th}$ rows represent how each feature is cascaded with the previous one.}
\label{fig:architecture}
\end{figure}

Next, we apply the adaptive average pooling operation on $\mathcal{F}$ to generate a one dimensional feature vector. This feature is passed through subsequent fully connected layers to generate the final prediction for $F_{W}$. See Fig.~\ref{fig:architecture} for a visualization of these procedures.

\begin{figure*}[t]
 \centering
 \includegraphics[width=0.99\textwidth]{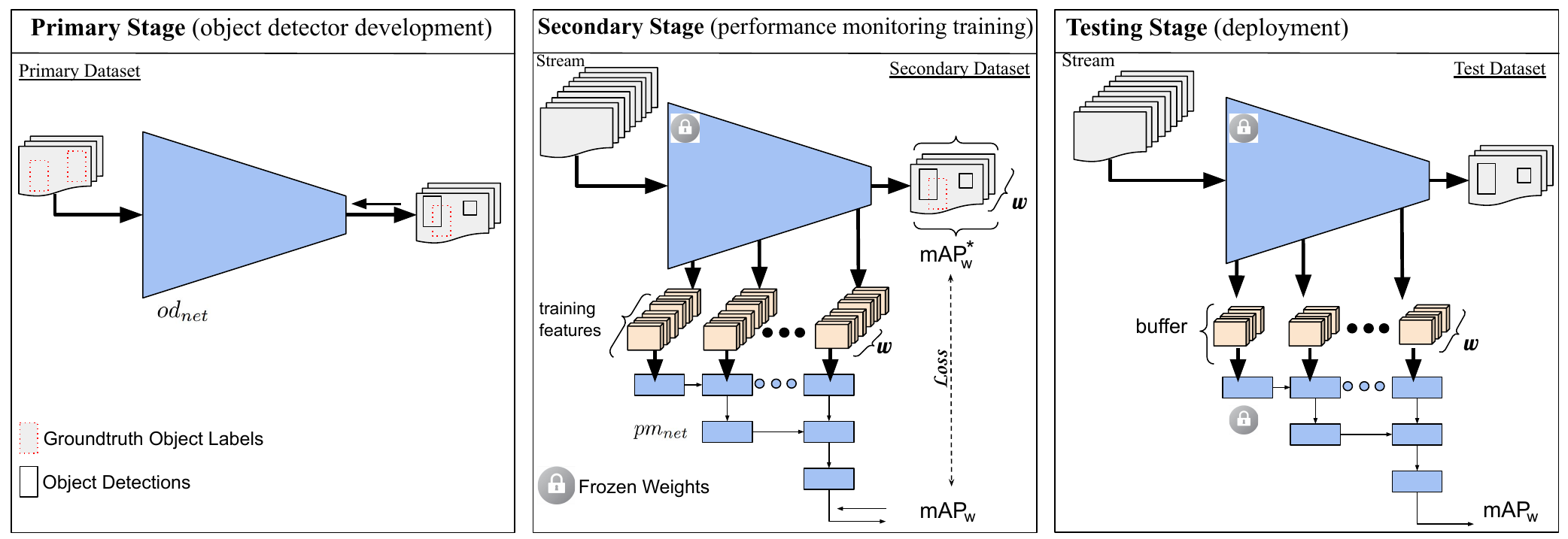}
 \caption{An overview of the three stages for training and testing our performance monitoring network ($pm_{net}$). First, the object detector is developed using a \textit{primary} training dataset. After the training is done, a \textit{secondary} training dataset consisting of video streams is used to collect inner features from the backbone network of the object detector and train $pm_{net}$ to predict the quality of the detector's per-window mAP. Finally, a third \textit{test} dataset of video streams is used to evaluate the performance monitoring network. The three datasets are independent of each other.}
\label{fig:overview}
\end{figure*}

\section{Experimental Setup}
\label{sec:experimental setup}
In this section, we will describe the settings that we used to evaluate our proposed approach.

\subsection{Experimental Steps}
We can describe the overall experimental procedure using three steps. At first, an object detector is trained using transfer learning techniques to detect different objects (vehicle,  pedestrian) from a dataset named as \textit{primary} training dataset. In the next step, the detector is used to detect similar objects from another dataset (\textit{secondary} training dataset) which was not used during the initial training phase. This \textit{secondary} training dataset consists of a stream of images. In this step, a sequential stream of images is fed to the object detector, and for each consecutive window of images, we collect the features for each window and calculate the corresponding mAP using the \textit{secondary} training dataset as ground-truth. Then, we train the proposed cascaded CNN, $pm_{net}$ to predict the mAP from the collected per-window features. Next, we use multiple metrics and another image stream dataset (\textit{test} dataset unused in previous steps to evaluate the proposed approach. See Fig.~\ref{fig:overview} for a high-level overview of these steps.

\subsection{Dataset}
We used multiple combinations of three different datasets (KITTI \cite{KITTI}, BDD \cite{yu2018bdd100k}, Waymo \cite{waymo}) to conduct all the experiments. In each setting, one dataset from KITTI and BDD has been used as \textit{primary} training dataset. Then we used one of the video datasets from KITTI, BDD and Waymo as the \textit{secondary} training dataset, which was not a part of \textit{primary} training dataset. To evaluate the system, we used one video dataset unused as \textit{primary} or \textit{secondary} training dataset. Each of the KITTI and BDD dataset has been split into $60\%$, $20\%$ and $20\%$ ratios to use as the \textit{primary}, \textit{secondary} and \textit{test} dataset. Besides, $50$ video segments of the Waymo dataset have been used as the \textit{test} dataset. The idea of \textit{primary} and \textit{secondary} datasets have been introduced to demonstrate the distribution shift between the training and deployment phase in the case of a driverless car. Moreover, the performance monitoring network has been trained using the \textit{secondary} dataset instead of \textit{primary} dataset following the frameworks proposed by \cite{zhang2014predicting} and \cite{ daftry2016introspective}.

\subsection{Training}
We trained two-stage Faster RCNN \cite{ren2015faster} and one-stage RetinaNet \cite{retinanet} object detection  networks  pre-trained on MS-COCO \cite{lin2014microsoft} dataset to detect vehicle and pedestrian from the KITTI and BDD dataset. Both of these networks use ResNet50 \cite{resnet} as their backbone.  To be interoperable among multiple datasets, classes like car, van, tram and bus have been assigned to vehicle class. Besides, pedestrian and person classes from all the datasets are denoted as the pedestrian class. Moreover, objects less than 25 pixels in width or height are removed from all the datasets. 

To generalize the object detection and performance monitoring network across multiple kinds of weather, lighting conditions and datasets -- we used the image augmentation library, Albumentations \cite{albu} with the default configuration to apply several augmentations such as random fog, snow and rain.  Table~\ref{tab:OD_mAP} shows the object detection accuracy in mAP for FRCNN and RetinaNet on multiple combination of \textit{primary} and \textit{secondary} dataset, respectively.

\begin{table}[b]
\caption{Object detection accuracy in mAP for FRCNN and RetinaNet network on multiple primary and secondary dataset for detecting vehicle and pedestrian.}
\label{tab:OD_mAP}
\centering
\begin{tabular}{llrr}
\hline
\multicolumn{2}{c}{Dataset} & \multicolumn{2}{c}{Object Detector} \\
Primary     & Secondary     & FRCNN          & RetinaNet          \\ \hline
kitti       & kitti         & 66.00          & 60.58              \\
kitti       & bdd           & 44.81          & 42.60              \\
kitti       & waymo         & 44.03          & 43.57              \\
bdd         & kitti         & 42.70          & 43.88              \\
bdd         & bdd           & 52.45          & 65.49              \\
bdd         & waymo         & 47.15          & 49.20              \\ \hline
\end{tabular}
\end{table}

We adopted the CORAL \cite{cao2019rank} framework that uses a set of binary classifiers to train $pm_{net}$ as an ordinal classifier. Each binary classifier predicts whether the per-window mAP is within a particular range. A decision threshold is used to control this prediction. The ordinal classifier has $5$ classes from $0$ to $4$, each incrementally spanning $0.2$ per-window mAP. In
this case, class $1$ is equivalent per-window mAP below $0.4$. In all of the experiments, $0.4$ is used as the critical threshold to train and evaluate the $pm_{net}$. Besides, $0.5$ intersection over union has been used to calculate the mAP. We used the Adam optimizer ~\cite{kingma2014adam} and an initial learning rate of $0.001$ with batch size $32$. 

\begin{figure}[]
\centering
\begin{subfigure}{0.65\columnwidth}
\centering
    \includegraphics[width=0.99\textwidth]{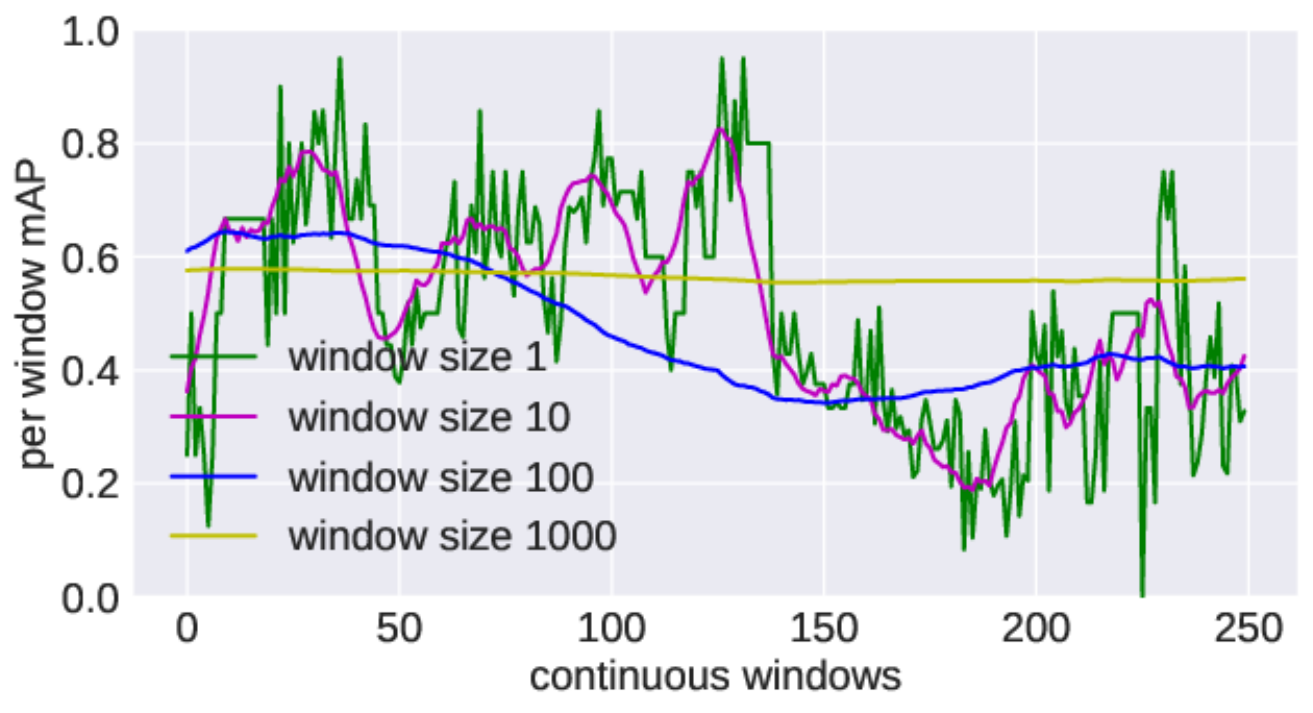}
    \caption{}
    \label{fig: mAP window size variation}
\end{subfigure}%
\begin{subfigure}{0.35\columnwidth}
\centering
    \includegraphics[width=0.99\textwidth]{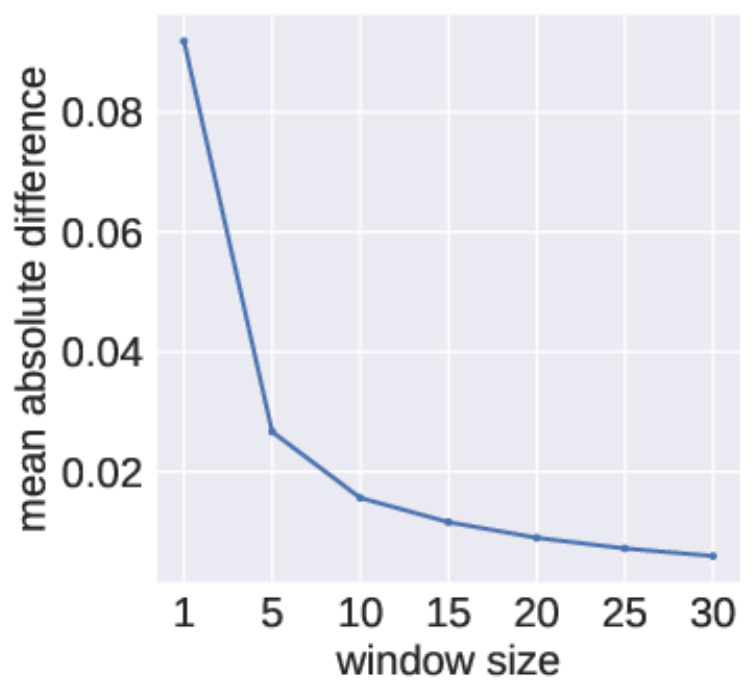}
    \caption{}
    \label{fig: mAP window size MAE}
\end{subfigure}
\caption{(a) Per-window mAP for multiple lengths of continuous windows. (b) Mean absolute difference between two consecutive per-window mAP for different length of continuous windows.}
\label{fig:window_size_variation}
\end{figure}

For all of the following experiments, we use a sliding window of $10$ frames. We empirically found that this value provides a balance between the high sensitivity of smaller windows and the smoothing effect of large windows, as shown in Fig.~\ref{fig:window_size_variation}.

\subsection{Evaluation Metrics}
We used mean absolute error (MAE), root mean squared error (RMSE), and zero-one error (ZOE) \cite{dembczynski2007ordinal}, which is the fraction of incorrect classification, as the evaluation metric for $pm_{net}$ ordinal classification task. To compare with the baselines and to evaluate how well the $pm_{net}$ can detect critical mAP label, we used true positive rate at 5\% false positive rate (TPR@FPR5), false positive rate at 95\% true positive rate (FPR@TPR95) and area under the ROC curve (AUROC) metric.

\subsection{Baseline Approaches}

\textbf{Baseline 1:} In \cite{Ramanagopal2018FailingTL}, Ramanagopal et. al. proposed an approach to identify perception failure of an object detection system. They used manually selected features like bounding box confidence and their mean and median overlap to identify false negative instances generated by an object detector. Following their approach, in this baseline, we extract a set of features from each image after performing the object detection. This set includes mean and median of all detected bounding box confidences, mean and median overlap, width and height of all detected bounding boxes on a normalized scale. We extracted these features from all images of each window and concatenated them together to generate a one-dimensional feature corresponding to the window. Next, each mAP per-window is converted into a binary label using a critical threshold of $0.4$. Any mAP lower than this threshold is assigned to the positive class; otherwise, the negative. Next, we train a fully connected binary classifier to predict the probability of each window to be assigned in the positive or negative class.

\textbf{Baseline 2:} In this baseline, we are using internal features from the last convolutional layer of a trained object detector backbone instead of handcrafted features. After each inference, we collect the $3D$ features from the final backbone layer and apply the average pooling technique to convert that $3D$ features into $1D$. After concatenating all $1D$ features from all the images of a window, we get a feature corresponding to that window. Using the critical threshold discussed in baseline 1, we assign each window into positive and negative classes. Then a fully connected binary classifier is trained to predict these classes from the window feature. 

We use class 1, which is equivalent to the critical threshold of $0.4$, to treat our ordinal classifier output as a binary classification. Consequently, classes from  $0$ to $1$ and $2$ to $4$ are assigned to positive and negative classes, respectively. This conversion allows us to compare our ordinal classifier with the binary classifier based baselines.

\section{Evaluation and Results }
\label{sec: evaluation and results}
In this section, we summarize how well the proposed performance monitoring network works as an ordinal classifier. Later, we evaluate the proposed network's accuracy to detect when the per-window mAP drops below the critical threshold of $0.4$.

\begin{table}[b]
\centering
\caption{MAE, RMSE and ZOE score for baselines and $pm_{net}$ trained using features collected from FRCNN.}
\label{tab:frcnn_accuracy}
\begin{tabular}{llllrrr}
\hline
Metric & \multicolumn{3}{c}{Datasets} & \multicolumn{1}{l}{Ours} & \multicolumn{1}{c}{Base-}  & \multicolumn{1}{c}{Base-}  \\
       & Primary  & Secondary  & Test & \multicolumn{1}{l}{}     & \multicolumn{1}{c}{line 2} & \multicolumn{1}{c}{line 1} \\ \hline
MAE$\downarrow$  & bdd   & kitti & waymo & \textbf{0.346} & 0.378 & 0.574 \\
     & bdd   & waymo & kitti & \textbf{0.181} & 0.375 & 0.409 \\
     & kitti & bdd   & waymo & \textbf{0.302} & 0.317 & 0.538 \\
     & kitti & waymo & bdd   & \textbf{0.241} & 0.316 & 0.446 \\ \hline
RMSE$\downarrow$ & bdd   & kitti & waymo & \textbf{0.594} & 0.611 & 0.819 \\
     & bdd   & waymo & kitti & \textbf{0.428} & 0.618 & 0.656 \\
     & kitti & bdd   & waymo & \textbf{0.569} & 0.595 & 0.783 \\
     & kitti & waymo & bdd   & \textbf{0.496} & 0.603 & 0.739 \\ \hline
ZOE$\downarrow$  & bdd   & kitti & waymo & \textbf{0.341} & 0.367 & 0.573 \\
     & bdd   & waymo & kitti & \textbf{0.182} & 0.356 & 0.386 \\
     & kitti & bdd   & waymo & \textbf{0.346} & 0.440 & 0.592 \\
     & kitti & waymo & bdd   & \textbf{0.269} & 0.369 & 0.513 \\ \hline
\end{tabular}
\end{table}

\textbf{Experiment 1:} Table~\ref{tab:frcnn_accuracy} shows the $pm_{net}$ ordinal classification accuracy for our proposed approach, Baseline 1 and Baseline 2 using MAE, RMSE and ZOE metric for four different dataset settings. Here, $od_{net}$ is trained using FRCNN  and  $pm_{net}$ is trained and evaluated using FRCNN backbone features. The second dataset setting for all metrics shows $pm_{net}$ error for ordinally classifying  $od_{net}$ performance on the KITTI \textit{test} dataset. $od_{net}$ and $pm_{net}$ are trained using \textit{primary} training dataset BDD and \textit{secondary} training dataset Waymo,  respectively. This setting demonstrates the lowest error in all the dataset settings.

\begin{table}[b]
\centering
\caption{MAE, RMSE and ZOE score for baselines and  $pm_{net}$ trained using features collected from RetinaNet.}
\label{tab:retinanet_accuracy}
\begin{tabular}{llllrrr}
\hline
\multicolumn{1}{c}{Metric} &
  \multicolumn{3}{c}{Datasets} &
  \multicolumn{1}{c}{Ours} &
  \multicolumn{1}{c}{Base-} &
  \multicolumn{1}{c}{Base-} \\
\multicolumn{1}{c}{} &
  \multicolumn{1}{c}{Primary} &
  \multicolumn{1}{c}{Secondary} &
  \multicolumn{1}{c}{Test} &
  \multicolumn{1}{c}{} &
  \multicolumn{1}{c}{line 2} &
  \multicolumn{1}{c}{line 1} \\ \hline
MAE$\downarrow$  & bdd   & kitti & waymo & \textbf{0.307} & 0.418 & 0.532 \\
     & bdd   & waymo & kitti & \textbf{0.272} & 0.387 & 0.487 \\
     & kitti & bdd   & waymo & \textbf{0.275} & 0.326 & 0.501 \\
     & kitti & waymo & bdd   & \textbf{0.290} & 0.433 & 0.501 \\ \hline
RMSE$\downarrow$ & bdd   & kitti & waymo & \textbf{0.601} & 0.675 & 0.833 \\
     & bdd   & waymo & kitti & \textbf{0.523} & 0.548 & 0.744 \\
     & kitti & bdd   & waymo & \textbf{0.554} & 0.639 & 0.771 \\
     & kitti & waymo & bdd   & \textbf{0.561} & 0.719 & 0.784 \\ \hline
ZOE$\downarrow$  & bdd   & kitti & waymo & \textbf{0.281} & 0.480 & 0.524 \\
     & bdd   & waymo & kitti & \textbf{0.259} & 0.390 & 0.498 \\
     & kitti & bdd   & waymo & \textbf{0.266} & 0.307 & 0.472 \\
     & kitti & waymo & bdd   & \textbf{0.277} & 0.429 & 0.486 \\ \hline
\end{tabular}
\end{table}

Table~\ref{tab:retinanet_accuracy} presents $pm_{net}$ error metric for similar dataset settings as Table~\ref{tab:frcnn_accuracy}. Here, $od_{net}$ is trained using RetinaNet object detection network and $pm_{net}$ is trained and evaluated using RetinaNet backbone features. In this table, \textit{primary} training dataset BDD, \textit{secondary} training dataset Waymo and \textit{test} dataset KITTI demonstrates the lowest error than other dataset settings. This observation is consistent with Table~\ref{tab:frcnn_accuracy} and suggests that the large diversity of BDD and Waymo dataset are effective for the performance monitoring of $pm_{net}$ in the KITTI dataset.

\textbf{Experiment 2:} This experiment compares the proposed performance monitoring network with the baselines. 

During the $pm_{net}$ evaluation, the decision threshold is varied from $0$ to $1$ to produce the $5$ class ordinal prediction for each threshold. Later, using the critical threshold, the ordinal class prediction is converted to a binary prediction to compute the TPR and FPR. Therefore, each decision threshold generates a pair of TPR, FPR and using these metrics; we calculate the TPR@FPR5, FPR@TPR95 and AUROC for our proposed approach.

As the two baselines use a binary classifier approach, we can use their predicted positive class probability and corresponding ground truth to calculate the  TPR@FPR5, FPR@TPR95 and AUROC metric.

\begin{table}[b]
\centering
\caption{$pm_{net}$ comparison with other baselines. Here, $pm_{net}$ is trained using FRCNN backbone features.}
\label{tab:frcnn_comparison}
\begin{tabular}{@{}llllrrr@{}}
\toprule
Metric & \multicolumn{3}{c}{Datasets} & Ours & \multicolumn{1}{c}{Base-}  & \multicolumn{1}{c}{Base-}  \\
       & Primary  & Secondary  & Test &      & \multicolumn{1}{c}{line 2} & \multicolumn{1}{c}{line 1} \\ \midrule
TPR@     & bdd   & kitti & waymo & \textbf{0.882} & 0.270 & 0.214 \\
FPR5$\uparrow$   & bdd   & waymo & kitti & \textbf{0.922} & 0.378 & 0.080 \\
             & kitti & bdd   & waymo & \textbf{0.916} & 0.320 & 0.157 \\
             & kitti & waymo & bdd   & \textbf{0.897} & 0.513 & 0.133 \\ \midrule
FPR@    & bdd   & kitti & waymo & \textbf{0.064} & 0.775 & 0.707 \\
TPR95$\downarrow$ & bdd   & waymo & kitti & \textbf{0.093} & 0.557 & 0.777 \\
             & kitti & bdd   & waymo & \textbf{0.133} & 0.766 & 0.827 \\
             & kitti & waymo & bdd   & \textbf{0.054} & 0.457 & 0.948 \\ \midrule
AUROC        & bdd   & kitti & waymo & \textbf{0.873} & 0.644 & 0.648 \\
$\uparrow$   & bdd   & waymo & kitti & \textbf{0.891} & 0.589 & 0.537 \\
             & kitti & bdd   & waymo & \textbf{0.892} & 0.610 & 0.573 \\
             & kitti & waymo & bdd   & \textbf{0.929} & 0.560 & 0.528 \\ \bottomrule
\end{tabular}
\end{table}

Table~\ref{tab:frcnn_comparison} shows the comparison between $pm_{net}$ and the two baselines using multiple metrics. For this table, the $od_{net}$ is trained using FRCNN and the $pm_{net}$ is trained and evaluated using FRCNN backbone features. In terms of TPR@FPR5, $pm_{net}$ outperforms both of the baselines. While the maximum TPR@FPR5 for $pm_{net}$ over four dataset settings is $0.922$, baseline 1 and 2 reach at maximum $0.214$ and $0.513$ respectively. For FPR@TPR95, the minimum score for our proposed approach is $0.054$. However, the minimum FPR@TPR95 for baseline 1 and 2 is $0.707$ and $0.457$, respectively. In AUROC metrics, $pm_{net}$ performs better than the baselines by obtaining $0.929$ while the maximum AUROC of both baselines is $0.610$. Although we have referred only to the maximum score of each individual metrics from the four dataset settings, our proposed approach outperforms the baselines in all metrics and dataset settings.

\begin{table}[b]
\centering
\caption{$pm_{net}$ comparison with other baselines. Here,  $pm_{net}$ is trained using  RetinaNet backbone features.}
\label{tab:retinanet comparison}
\begin{tabular}{@{}llllrrr@{}}
\toprule
Metric & \multicolumn{3}{c}{Datasets} & Ours & \multicolumn{1}{c}{Base-}  & \multicolumn{1}{c}{Base-}  \\
       & Primary  & Secondary  & Test &      & \multicolumn{1}{c}{line 2} & \multicolumn{1}{c}{line 1} \\ \midrule
TPR@   & bdd   & kitti & waymo & \textbf{0.875} & 0.228 & 0.157 \\
FPR5$\uparrow$ & bdd   & waymo & kitti & \textbf{0.953} & 0.380 & 0.080 \\
           & kitti & bdd   & waymo & \textbf{0.915} & 0.576 & 0.214 \\
           & kitti & waymo & bdd   & \textbf{0.889} & 0.708 & 0.133 \\ \midrule
FPR@  & bdd   & kitti & waymo & \textbf{0.098} & 0.576 & 0.827 \\
TPR95$\downarrow$ & bdd   & waymo & kitti & \textbf{0.109} & 0.765 & 0.777 \\
           & kitti & bdd   & waymo & \textbf{0.053} & 0.479 & 0.707 \\
           & kitti & waymo & bdd   & \textbf{0.076} & 0.911 & 0.948 \\ \midrule
AUROC      & bdd   & kitti & waymo & \textbf{0.860} & 0.567 & 0.573 \\
$\uparrow$ & bdd   & waymo & kitti & \textbf{0.917} & 0.596 & 0.537 \\
           & kitti & bdd   & waymo & \textbf{0.915} & 0.564 & 0.648 \\
           & kitti & waymo & bdd   & \textbf{0.873} & 0.746 & 0.528 \\ \bottomrule
\end{tabular}
\end{table}

\begin{figure}[]
\centering
\centering
    \includegraphics[width=0.99\columnwidth]{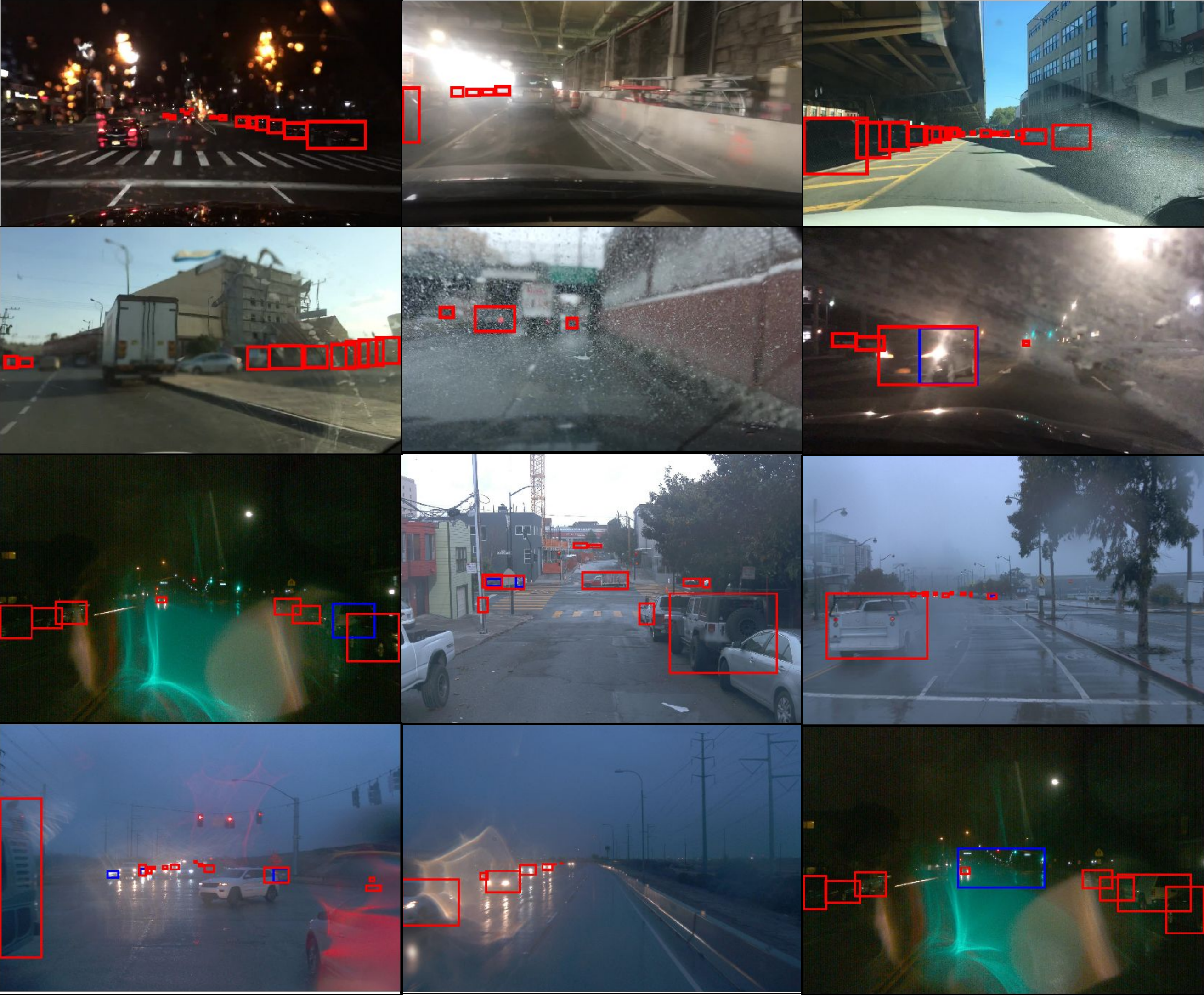}
\caption{Example of performance drop encountered by the object detection system. Red and Blue coloured boxes represent the false negative and false positive error made by the object detector in each input frame. $1^{st}$ and $2^{nd}$ rows represent Faster RCNN and RetinaNet detection performance for primary dataset KITTI and secondary dataset BDD. $3^{rd}$ and $4^{th}$ rows show Faster RCNN and RetinaNet performance for primary dataset BDD and secondary dataset Waymo. In all cases, the mAP is lower than $0.4$, and detection confidence is higher than $0.5$.}
\label{fig:samples}
\end{figure}

Table~\ref{tab:retinanet comparison} represents the comparative accuracy among $pm_{net}$ and two other baselines. Here, the underlying object detector is trained using the RetinaNet network and the corresponding performance monitoring network is trained and evaluated using features collected from the RetinaNet backbone. In this case, for the TPR@FPR5 metric, the maximum score that our proposed approach achieves out of four dataset settings is $0.953$ while the maximum of two baselines among these four settings is $0.708$. In the case of FPT@TPR95 and AUROC, our proposed approach outperforms both of the baselines. Fig.~\ref{fig:samples} demonstrates the multiple example of performance drop when the object detector is trained and tested on \textit{primary} and \textit{secondary} dataset, respectively. In all cases, the mAP is lower than the critical threshold, and our performance monitoring network flags them correctly.

This experimental result demonstrates the effectiveness of internal features of an object detector for performance monitoring task. Although Baseline 1 and Baseline 2 can be used for performance monitoring, our proposed approach outperforms them because of the cascaded architecture's internal feature usage. Here, the cascaded architecture can capture the gradual change in the per-window mAP better than the baselines. Moreover, instead of using features only from the last layer of the object detection backbone, our proposed approach extracts internal features from all convolutional layers. This approach derives global and local multi-level semantic features for better performance monitoring. 

\textbf{Experiment 3:} In order to monitor object detection performance online, we are required to simultaneously use the performance monitoring network along with the object detection system. Hence, the inference time and GPU memory requirement of $pm_{net}$ should be minimal for practical usage. On average $pm_{net}$ and $od_{net}$ inference time is $3.34 \pm 0.126$ ms and $28.11 \pm 0.404$ ms in our TITAN V GPU workstation. Besides, $pm_{net}$ uses $243$ MB of GPU memory which is $20.81\%$ of memory used by the $od_{net}$.

\section{Conclusion}
\label{sec:conclusion}
As deep learning-based object detection becomes essential components of a wide variety of robotic systems, the ability to continuously assess and monitor their performance during the deployment phase becomes critical to ensure the safety and reliability of the whole system. In this paper, we proposed a specialized performance monitoring network that can predict the quality of the mAP of the object detector, which can be used to inform downstream components in the robotic system about the expected object detection reliability. We show the effectiveness of our approach using a combination of different autonomous driving datasets and object detectors.

\bibliographystyle{IEEEtran}
\bibliography{ref}

\begin{thebibliography}{10}
\providecommand{\url}[1]{#1}
\csname url@rmstyle\endcsname
\providecommand{\newblock}{\relax}
\providecommand{\bibinfo}[2]{#2}
\providecommand\BIBentrySTDinterwordspacing{\spaceskip=0pt\relax}
\providecommand\BIBentryALTinterwordstretchfactor{4}
\providecommand\BIBentryALTinterwordspacing{\spaceskip=\fontdimen2\font plus
\BIBentryALTinterwordstretchfactor\fontdimen3\font minus
  \fontdimen4\font\relax}
\providecommand\BIBforeignlanguage[2]{{%
\expandafter\ifx\csname l@#1\endcsname\relax
\typeout{** WARNING: IEEEtran.bst: No hyphenation pattern has been}%
\typeout{** loaded for the language `#1'. Using the pattern for}%
\typeout{** the default language instead.}%
\else
\language=\csname l@#1\endcsname
\fi
#2}}

\bibitem{tian2019fcos}
Z.~Tian, C.~Shen, H.~Chen, and T.~He, ``{FCOS: Fully Convolutional One-stage
  Object Detection},'' in \emph{Proceedings of the IEEE international
  conference on computer vision}, 2019, pp. 9627--9636.

\bibitem{ren2015faster}
S.~Ren, K.~He, R.~Girshick, and J.~Sun, ``{Faster R-CNN: Towards Real-time
  Object Detection With Region Proposal Networks},'' in \emph{Advances in
  neural information processing systems}, 2015, pp. 91--99.

\bibitem{liu2016ssd}
W.~Liu, D.~Anguelov, D.~Erhan, C.~Szegedy, S.~Reed, C.-Y. Fu, and A.~C. Berg,
  ``{SSD: Single Shot Multibox Detector},'' in \emph{European conference on
  computer vision}.\hskip 1em plus 0.5em minus 0.4em\relax Springer, 2016, pp.
  21--37.

\bibitem{Duan2019CenterNetKT}
K.~Duan, S.~Bai, L.~Xie, H.~Qi, Q.~Huang, and Q.~Tian, ``{CenterNet: Keypoint
  Triplets for Object Detection},'' \emph{2019 IEEE/CVF International
  Conference on Computer Vision (ICCV)}, pp. 6568--6577, 2019.

\bibitem{Bochkovskiy2020YOLOv4OS}
A.~Bochkovskiy, C.-Y. Wang, and H.-Y.~M. Liao, ``{YOLOv4: Optimal Speed and
  Accuracy of Object Detection},'' \emph{ArXiv}, vol. abs/2004.10934, 2020.

\bibitem{He2019RethinkingIP}
K.~He, R.~B. Girshick, and P.~Doll{\'a}r, ``{Rethinking ImageNet
  Pre-Training},'' \emph{2019 IEEE/CVF International Conference on Computer
  Vision (ICCV)}, pp. 4917--4926, 2019.

\bibitem{Cai2018CascadeRD}
Z.~Cai and N.~Vasconcelos, ``{Cascade R-CNN: Delving Into High Quality Object
  Detection},'' \emph{2018 IEEE/CVF Conference on Computer Vision and Pattern
  Recognition}, pp. 6154--6162, 2018.

\bibitem{Li2018DetNetAB}
Z.~Li, C.~Peng, G.~Yu, X.~Zhang, Y.~Deng, and J.~Sun, ``{DetNet: A Backbone
  network for Object Detection},'' \emph{ArXiv}, vol. abs/1804.06215, 2018.

\bibitem{Lin2017FocalLF}
T.-Y. Lin, P.~Goyal, R.~B. Girshick, K.~He, and P.~Doll{\'a}r, ``{Focal Loss
  for Dense Object Detection},'' \emph{2017 IEEE International Conference on
  Computer Vision (ICCV)}, pp. 2999--3007, 2017.

\bibitem{Carion2020EndtoEndOD}
N.~Carion, F.~Massa, G.~Synnaeve, N.~Usunier, A.~M. Kirillov, and S.~Zagoruyko,
  ``{End-to-End Object Detection with Transformers},'' \emph{ArXiv}, vol.
  abs/2005.12872, 2020.

\bibitem{morris2007robotic}
A.~C. Morris, ``{Robotic Introspection for Exploration and Mapping of
  Subterranean Environments},'' Ph.D. dissertation, Carnegie Mellon University,
  The Robotics Institute, 2007.

\bibitem{grimmett2016introspective}
H.~Grimmett, R.~Triebel, R.~Paul, and I.~Posner, ``{Introspective
  Classification for Robot Perception},'' \emph{The International Journal of
  Robotics Research}, vol.~35, no.~7, pp. 743--762, 2016.

\bibitem{Triebel2013DrivenLF}
R.~Triebel, H.~Grimmett, R.~Paul, and I.~Posner, ``{Driven Learning for
  Driving: How Introspection Improves Semantic Mapping},'' in \emph{ISRR},
  2013.

\bibitem{zhang2014predicting}
P.~Zhang, J.~Wang, A.~Farhadi, M.~Hebert, and D.~Parikh, ``{Predicting Failures
  of Vision Systems},'' in \emph{Proceedings of the IEEE Conference on Computer
  Vision and Pattern Recognition}, 2014, pp. 3566--3573.

\bibitem{daftry2016introspective}
S.~Daftry, S.~Zeng, J.~A. Bagnell, and M.~Hebert, ``{Introspective Perception:
  Learning to Predict Failures in Vision Systems},'' in \emph{2016 IEEE/RSJ
  International Conference on Intelligent Robots and Systems (IROS)}.\hskip 1em
  plus 0.5em minus 0.4em\relax IEEE, 2016, pp. 1743--1750.

\bibitem{wang2018towards}
P.~Wang and N.~Vasconcelos, ``{Towards Realistic Predictors},'' in
  \emph{Proceedings of the European Conference on Computer Vision (ECCV)},
  2018, pp. 36--51.

\bibitem{Gurau2018LearnFE}
C.~Gurau, D.~Rao, C.~H. Tong, and I.~Posner, ``{Learn From Experience:
  Probabilistic Prediction of Perception Performance to Avoid Failure},''
  \emph{The International Journal of Robotics Research}, vol.~37, pp. 981 --
  995, 2018.

\bibitem{Jiang2018ToTO}
H.~Jiang, B.~Kim, and M.~R. Gupta, ``{To Trust or Not to Trust A Classifier},''
  in \emph{NeurIPS}, 2018.

\bibitem{Hendrycks2017ABF}
D.~Hendrycks and K.~Gimpel, ``{A Baseline for Detecting Misclassified and
  Out-of-Distribution Examples in Neural Networks},'' \emph{ArXiv}, vol.
  abs/1610.02136, 2017.

\bibitem{corbiere2019addressing}
C.~Corbi{\`e}re, N.~Thome, A.~Bar-Hen, M.~Cord, and P.~P{\'e}rez, ``{Addressing
  Failure Prediction by Learning Model Confidence},'' in \emph{Advances in
  Neural Information Processing Systems}, 2019, pp. 2902--2913.

\bibitem{Gal2016DropoutAA}
Y.~Gal and Z.~Ghahramani, ``{Dropout as a Bayesian Approximation: Representing
  Model Uncertainty in Deep Learning},'' in \emph{ICML}, 2016.

\bibitem{Devries2018LeveragingUE}
T.~Devries and G.~W. Taylor, ``{Leveraging Uncertainty Estimates for Predicting
  Segmentation Quality},'' \emph{ArXiv}, vol. 1807.00502, 2018.

\bibitem{Huang2018EfficientUE}
P.-Y. Huang, W.~T. Hsu, C.-Y. Chiu, T.-F. Wu, and M.~Sun, ``{Efficient
  Uncertainty Estimation for Semantic Segmentation in Videos},'' in
  \emph{ECCV}, 2018.

\bibitem{hendrycks2016baseline}
D.~Hendrycks and K.~Gimpel, ``A baseline for detecting misclassified and
  out-of-distribution examples in neural networks,'' \emph{arXiv preprint
  arXiv:1610.02136}, 2016.

\bibitem{liang2017enhancing}
S.~Liang, Y.~Li, and R.~Srikant, ``{Enhancing the Reliability of
  Out-of-Distribution Image Detection in Neural Networks},'' \emph{arXiv
  preprint arXiv:1706.02690}, 2017.

\bibitem{hendrycks2018deep}
D.~Hendrycks, M.~Mazeika, and T.~Dietterich, ``{Deep anomaly detection with
  outlier exposure},'' \emph{arXiv preprint arXiv:1812.04606}, 2018.

\bibitem{lee2017training}
K.~Lee, H.~Lee, K.~Lee, and J.~Shin, ``Training confidence-calibrated
  classifiers for detecting out-of-distribution samples,'' \emph{arXiv preprint
  arXiv:1711.09325}, 2017.

\bibitem{lee2018simple}
K.~Lee, K.~Lee, H.~Lee, and J.~Shin, ``A simple unified framework for detecting
  out-of-distribution samples and adversarial attacks,'' \emph{arXiv preprint
  arXiv:1807.03888}, 2018.

\bibitem{Miller2018DropoutSF}
D.~Miller, L.~Nicholson, F.~Dayoub, and N.~S{\"u}nderhauf, ``{Dropout Sampling
  for Robust Object Detection in Open-Set Conditions},'' \emph{2018 IEEE
  International Conference on Robotics and Automation (ICRA)}, pp. 1--7, 2018.

\bibitem{Cheng2018DecoupledCR}
B.~Cheng, Y.~Wei, H.~Shi, R.~S. Feris, J.~Xiong, and T.~S. Huang, ``{Decoupled
  Classification Refinement: Hard False Positive Suppression for Object
  Detection},'' \emph{ArXiv}, vol. abs/1810.04002, 2018.

\bibitem{Rahman2019DidYM}
Q.~M. Rahman, N.~S{\"u}nderhauf, and F.~Dayoub, ``{Did You Miss the Sign? A
  False Negative Alarm System for Traffic Sign Detectors},'' \emph{2019
  IEEE/RSJ International Conference on Intelligent Robots and Systems (IROS)},
  pp. 3748--3753, 2019.

\bibitem{Ramanagopal2018FailingTL}
M.~S. Ramanagopal, C.~Anderson, R.~Vasudevan, and M.~Johnson-Roberson,
  ``{Failing to Learn: Autonomously Identifying Perception Failures for
  Self-Driving Cars},'' \emph{IEEE Robotics and Automation Letters}, vol.~3,
  pp. 3860--3867, 2018.

\bibitem{azulay2018deep}
A.~Azulay and Y.~Weiss, ``{Why Do Deep Convolutional Networks Generalize So
  Poorly to Small Image Transformations?}'' \emph{arXiv preprint
  arXiv:1805.12177}, 2018.

\bibitem{cardoso2007learning}
J.~S. Cardoso and J.~F. Costa, ``{ Learning to Classify Ordinal Data: The Data
  Replication Method},'' \emph{Journal of Machine Learning Research}, vol.~8,
  no. Jul, pp. 1393--1429, 2007.

\bibitem{KITTI}
A.~Geiger, P.~Lenz, and R.~Urtasun, ``{Are We Ready for Autonomous Driving? The
  KITTI Vision Benchmark Suite},'' in \emph{Conference on Computer Vision and
  Pattern Recognition (CVPR)}, 2012.

\bibitem{yu2018bdd100k}
F.~Yu, W.~Xian, Y.~Chen, F.~Liu, M.~Liao, V.~Madhavan, and T.~Darrell,
  ``{BDD100k: A Diverse Driving Video Database With Scalable Annotation
  Tooling},'' \emph{arXiv preprint arXiv:1805.04687}, vol.~2, no.~5, p.~6,
  2018.

\bibitem{waymo}
P.~Sun, H.~Kretzschmar, X.~Dotiwalla, A.~Chouard, V.~Patnaik, P.~Tsui, J.~Guo,
  Y.~Zhou, Y.~Chai, B.~Caine, V.~Vasudevan, W.~Han, J.~Ngiam, H.~Zhao,
  A.~Timofeev, S.~Ettinger, M.~Krivokon, A.~Gao, A.~Joshi, Y.~Zhang, J.~Shlens,
  Z.~Chen, and D.~Anguelov, ``{Scalability in Perception for Autonomous
  Driving: Waymo Open Dataset},'' in \emph{Proceedings of the IEEE/CVF
  Conference on Computer Vision and Pattern Recognition (CVPR)}, June 2020.

\bibitem{retinanet}
T.-Y. Lin, P.~Goyal, R.~Girshick, K.~He, and P.~Doll{\'a}r, ``{Focal Loss for
  Dense Object Detection},'' in \emph{Proceedings of the IEEE international
  conference on computer vision}, 2017, pp. 2980--2988.

\bibitem{lin2014microsoft}
T.-Y. Lin, M.~Maire, S.~Belongie, J.~Hays, P.~Perona, D.~Ramanan,
  P.~Doll{\'a}r, and C.~L. Zitnick, ``{Microsoft COCO: Common Objects in
  Context},'' in \emph{European conference on computer vision}.\hskip 1em plus
  0.5em minus 0.4em\relax Springer, 2014, pp. 740--755.

\bibitem{resnet}
K.~He, X.~Zhang, S.~Ren, and J.~Sun, ``{Deep Residual Learning for Image
  Recognition},'' in \emph{Proceedings of the IEEE conference on computer
  vision and pattern recognition}, 2016, pp. 770--778.

\bibitem{albu}
\BIBentryALTinterwordspacing
A.~Buslaev, V.~I. Iglovikov, E.~Khvedchenya, A.~Parinov, M.~Druzhinin, and
  A.~A. Kalinin, ``Albumentations: Fast and flexible image augmentations,''
  \emph{Information}, vol.~11, no.~2, 2020. [Online]. Available:
  \url{https://www.mdpi.com/2078-2489/11/2/125}
\BIBentrySTDinterwordspacing

\bibitem{cao2019rank}
W.~Cao, V.~Mirjalili, and S.~Raschka, ``{Rank-consistent Ordinal Regression for
  Neural Networks},'' \emph{arXiv preprint arXiv:1901.07884}, 2019.

\bibitem{kingma2014adam}
D.~P. Kingma and J.~Ba, ``{Adam: A Method for Stochastic Optimization},''
  \emph{arXiv preprint arXiv:1412.6980}, 2014.

\bibitem{dembczynski2007ordinal}
K.~Dembczy{\'n}ski, W.~Kot{\l}owski, and R.~S{\l}owi{\'n}ski, ``{Ordinal
  Classification with Decision Rules},'' in \emph{International Workshop on
  Mining Complex Data}.\hskip 1em plus 0.5em minus 0.4em\relax Springer, 2007,
  pp. 169--181.

\end{thebibliography}
\end{document}